\documentclass[10pt,twocolumn,letterpaper]{article}

\usepackage[pagenumbers]{cvpr}      %

\usepackage[dvipsnames]{xcolor}

\usepackage{enumitem}
\usepackage{cuted}
\usepackage{arydshln}
\usepackage{pifont}

\definecolor{cvprblue}{rgb}{0.21,0.49,0.74}
\usepackage[pagebackref,breaklinks,colorlinks,citecolor=cvprblue]{hyperref}

\def\paperID{2903} %
\def\confName{CVPR}
\def\confYear{2024}

\title{Tactile-Augmented Radiance Fields}

\author{
Yiming Dou\textsuperscript{1} \quad
Fengyu Yang\textsuperscript{2} \quad
Yi Liu\textsuperscript{1} \quad
Antonio Loquercio\textsuperscript{3} \quad
Andrew Owens\textsuperscript{1} \quad
\vspace{3mm} \\
\textsuperscript{1}University of Michigan \quad \textsuperscript{2}Yale University
\quad \textsuperscript{3}UC Berkeley  %
\\
}

\newcommand{\mysection}[1]{\vspace{-1.5mm}\section{#1}}

\newcommand{\mypar}[1]{\vspace{1mm}\noindent {\bf #1}~~}

\newcommand{\im}[0]{\mathbf v}
\newcommand{\depth}[0]{\mathbf d}
\newcommand{\background}[0]{\mathbf b}
\newcommand{\touch}[0]{\boldsymbol{\tau}}

\newcommand{\bu}[0]{\mathbf u}

\newcommand{\touchparams}{\phi}
\newcommand{\nerfparams}{\theta}

\newcommand{\fieldt}[0]{F_{\touchparams}}
\newcommand{\fieldv}[0]{F_{\nerfparams}}

\newcommand{\pt}[0]{\mathbf x}

\newcommand{\view}[0]{\mathbf r}
\newcommand{\rgb}[0]{\mathbf c}

\newcommand{\vtfield}[0]{TaRF\xspace}

\newcommand{\vidt}[0]{\mathbf{v}}

\usepackage{bbding}
\usepackage{caption}  %
\usepackage{subcaption}  %
\usepackage{booktabs}  %
\usepackage{tikz}
\usepackage[accsupp]{axessibility}

\usepackage{tabularx}

\begin{document}
\maketitle
\begin{strip}
\centering
\vspace{-10mm}
    \centering
    \raggedright
    \vspace{-4mm} %
   \includegraphics[width=\textwidth]{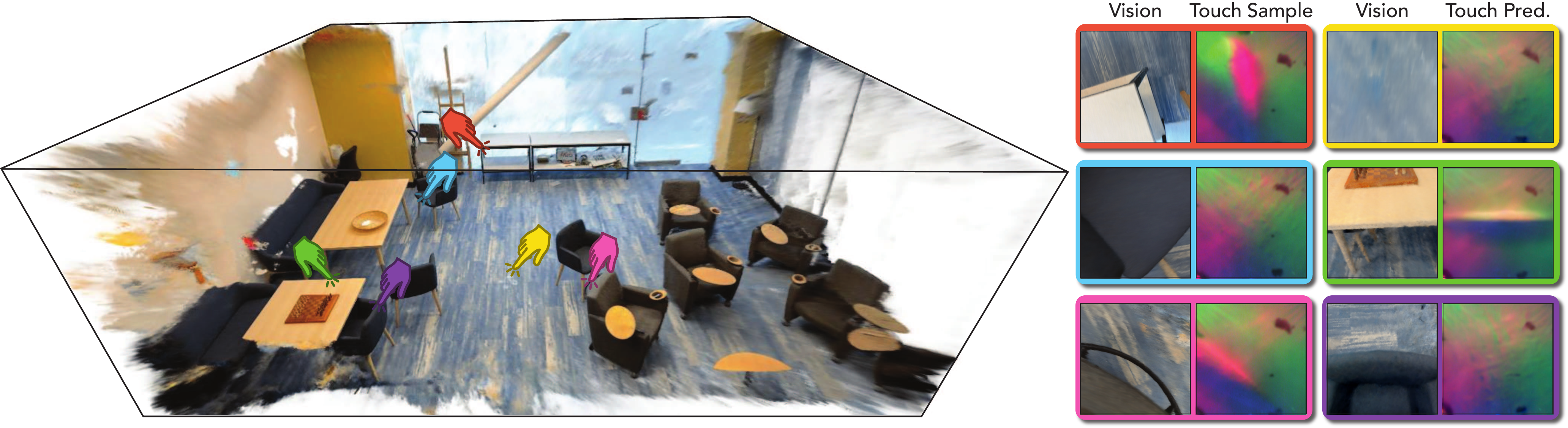}
    \vspace{0.25mm}
    \captionof{figure}{
        \textbf{Tactile-augmented radiance fields.} 
        We capture a {\em tactile-augmented radiance field} (\vtfield) from photos and sparsely sampled touch probes. To do this, we register the captured visual and tactile signals into a shared 3D space, then train a diffusion model to impute touch at other locations within the scene. Here, we visualize two touch probes and their (color coded) 3D positions in the scene. We also show two touch signals estimated by the diffusion model. The touch signals were collected using a vision-based touch sensor~\cite{lambeta2020digit} that represents the touch signals as images. Please see our \href{https://dou-yiming.github.io/TaRF}{project page} for video results. %
    }   
\label{fig:teaser}
\end{strip}

\begin{abstract}
\vspace{-2mm}

We present a scene representation, which we call a {\em tactile-augmented radiance field} (TaRF), that brings vision and touch into a shared 3D space. 
This representation can be used to estimate the visual and tactile signals for a given 3D position within a scene.
We capture a scene's TaRF from a collection of photos and sparsely sampled touch probes.
Our approach makes use of two insights: (i) common vision-based touch sensors are built on ordinary cameras and thus can be registered to images using methods from multi-view geometry, and (ii) visually and structurally similar regions of a scene share the same tactile features.
We use these insights to register touch signals to a captured visual scene, and to train a conditional diffusion model that, provided with an RGB-D image rendered from a neural radiance field, generates its corresponding tactile signal.
To evaluate our approach, we collect a dataset of TaRFs. This dataset contains more touch samples than previous real-world datasets, and it provides spatially aligned visual signals for each captured touch signal. We demonstrate the accuracy of our cross-modal generative model and the utility of the captured visual-tactile data on several downstream tasks.
Project page:  \small{\url{https://dou-yiming.github.io/TaRF}}.

\end{abstract}
    
\section{Introduction}
\label{sec:intro}

As humans, our ability to perceive the world relies crucially on cross-modal associations between sight and touch~\cite{smith2005development,graziano1995representation}. Tactile sensing provides a detailed understanding of material properties and microgeometry, such as the intricate patterns of bumps on rough surfaces and the complex motions that soft objects make when they deform.
This type of understanding, which largely eludes today's computer vision models, is a critical component of applications that require reasoning about physical contact, such as robotic locomotion~\cite{loquercio2023learning,Li2023ViHOPE, margolis2023learning,hoepflinger2010haptic,kolvenbach2019haptic,bednarek2019touching} and manipulation~\cite{calandra2017feeling,calandra2018more,qi2023general,yin2023rotating,church2020deep}, and methods that simulate the behavior of materials~\cite{purushwalkam2019bounce,owens2016visually,bouman2013estimating,davis2015image}.

\begin{table}[t]
\begin{minipage}[b]{\linewidth}
\centering
\resizebox{\columnwidth}{!}{
    \centering
    \begin{tabular}{@{}llcll}
         \toprule
                                 Dataset   & {Samples} & {Aligned} & {Scenario} & {Source}\\ 
         \midrule
        More Than a Feeling~\cite{calandra2018more} & 6.5k  & \textcolor{red}{\ding{53}} & Tabletop & Robot\\
        Feeling of Success~\cite{calandra2017feeling}  & 9.3k &\textcolor{red}{\ding{53}} & Tabletop & Robot\\
        VisGel~\cite{li2019connecting}  & 12k&\textcolor{red}{\ding{53}}  & Tabletop & Robot\\
        SSVTP~\cite{kerr2022ssvtp} & 4.6k &\textcolor{green}{\ding{51}} & Tabletop & Robot\\
        ObjectFolder 1.0~\cite{gao2021objectfolder}  & -- & \textcolor{green}{\ding{51}} & Object & Synthetic\\
        ObjectFolder 2.0~\cite{gao2022objectfolder}  & -- & \textcolor{green}{\ding{51}} & Object & Synthetic\\

        ObjectFolder Real~\cite{gao2023objectfolder}  & 3.7k  &\textcolor{red}{\ding{53}} & Object & Robot\\
        Burka et al.~\cite{burka2018instrumentation}  & 1.1k  &\textcolor{red}{\ding{53}} & Sub-scene & Human\\
        Touch and Go~\cite{yang2022touch} &  13.9k  & \textcolor{red}{\ding{53}} & Sub-scene &  Human\\ 
        YCB-Slide$^*$~\cite{suresh2023midastouch} & -  &\textcolor{green}{\ding{51}} & Object & Human\\
        Touching a NeRF~\cite{zhong2023touching} & 1.2k  & \textcolor{green}{\ding{51}} & Object & Robot\\
\cdashline{1-5}
        {\bf TaRF (Ours)} & \textbf{19.3k} & \textcolor{green}{\ding{51}} & \textbf{Full scene} & Human \\
        \bottomrule
    \end{tabular}}
    \end{minipage}
    \vspace{-3mm}
    \caption{{\bf Dataset comparison}. We present the number of real visual-tactile pairs and whether such pairs are visually aligned, i.e., whether the visual image includes an occlusion-free view of the touched surface. $^{*}$YCB-Slide has real-world touch probes but \emph{synthetic} images rendered with CAD models of YCB objects on a white background~\cite{calli2015ycb}.} \vspace{-4mm}
    \label{tab:dataset_comp} %
\end{table}

\begin{figure}[t]
    \centering
    \newlength{\minipagewidth}
    \setlength{\minipagewidth}{0.075\textwidth} %
    \addtolength{\minipagewidth}{-0.0001\textwidth} %
    \captionsetup[sub]{labelformat=empty}
    \begin{subfigure}[t]{\minipagewidth}
        \centering
        \captionsetup{font=scriptsize}
        \includegraphics[width=\minipagewidth]{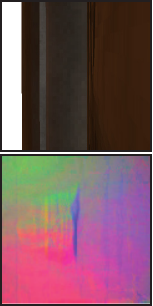}
        \caption{OF 2.0~\cite{gao2022objectfolder}}
    \end{subfigure}\hfill
    \begin{subfigure}[t]{\minipagewidth}
        \centering
        \captionsetup{font=scriptsize}
        \includegraphics[width=\minipagewidth]{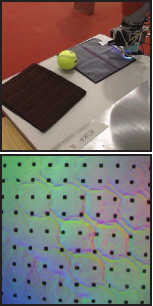}
        \caption{VisGel~\cite{li2019connecting}}
    \end{subfigure}\hfill
    \begin{subfigure}[t]{\minipagewidth}
        \centering
        \captionsetup{font=scriptsize}
        \includegraphics[width=\minipagewidth]{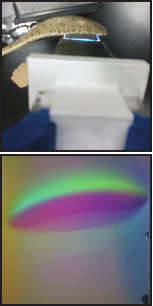}
        \caption{OF Real~\cite{gao2023objectfolder}}
    \end{subfigure}\hfill
    \begin{subfigure}[t]{\minipagewidth}
        \centering
        \captionsetup{font=scriptsize}
        \includegraphics[width=\minipagewidth]{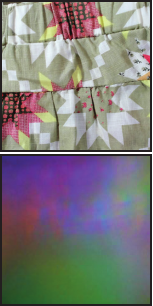}
        \caption{SSVTP~\cite{kerr2022ssvtp}}
    \end{subfigure}\hfill
    \begin{subfigure}[t]{\minipagewidth}
        \centering
        \captionsetup{font=scriptsize}
        \includegraphics[width=\minipagewidth]{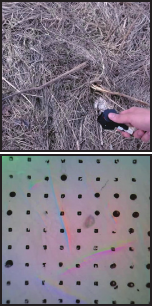}
            \caption{TG~\cite{yang2022touch}}
    \end{subfigure}\hfill
    \begin{subfigure}[t]{\minipagewidth}
        \centering
        \captionsetup{font=scriptsize}
        \includegraphics[width=\minipagewidth]{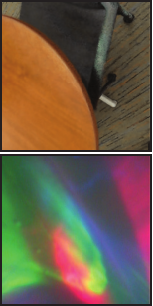}
        \caption{TaRF (Ours)}
    \end{subfigure}\hfill
    \vspace{-2mm}
    \caption{{\bf Visual-tactile examples.} In contrast to the visual-tactile data captured in previous work, our approach allows us to sample unobstructed images that are spatially aligned with the touch signal, from arbitrary 3D viewpoints using a NeRF.} \vspace{-4mm}
    \label{fig:dataset_comp}
\end{figure}

In comparison to many other modalities, collecting tactile data is an expensive and tedious process, since it requires direct physical interaction with the environment.
A recent line of work has addressed this problem by having humans or robots probe the environment with touch sensors (see Table~\ref{tab:dataset_comp}). Early efforts have been focused on capturing the properties of only a few objects either in simulation~\cite{gao2021objectfolder,gao2022objectfolder,suresh2023midastouch} or in lab-controlled settings~\cite{gao2023objectfolder, calandra2017feeling, calandra2018more,kerr2022ssvtp,li2019connecting,suresh2023midastouch,zhong2023touching}, which may not fully convey the diversity of tactile signals in natural environments. Other works have gone beyond a lab setting and have collected touch from real scenes~\cite{burka2018instrumentation,yang2022touch}. 
However, existing datasets lack aligned visual and tactile information, since the touch sensor and the person (or robot) that holds it often occlude large portions of the visual scene (Fig.~\ref{fig:dataset_comp}). 
These datasets also contain only a sparse set of touch signals for each scene, and it is not clear how the sampled touch signals relate to each other in 3D. %

\begin{figure}[t]
    \centering
    \includegraphics[width=0.47\textwidth]{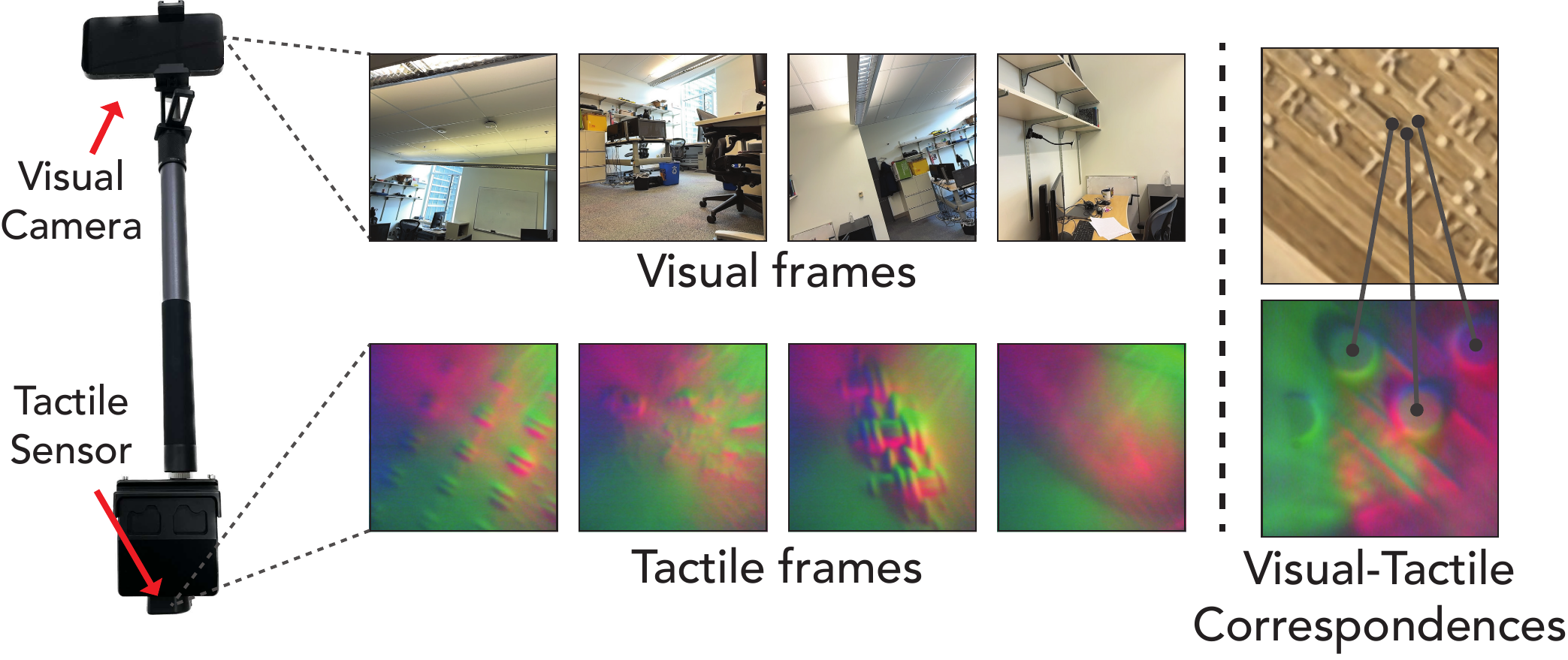}
            \caption{{\bf Capturing setup.} (a) We record paired vision and touch signals using a camera attached to a touch sensor. (b) We estimate the relative pose between the touch sensor and the camera using correspondences between sight and touch.}
    \label{fig:capturing_device}
\end{figure}

In this work, we present a simple and low-cost procedure to capture {quasi-dense, scene-level, and spatially-aligned visual and touch data} (Fig.~\ref{fig:teaser}). We call the resulting scene representation a {\em tactile-augmented radiance field} (\vtfield).
We remove the need for robotic collection by leveraging a 3D scene representation (a NeRF~\cite{mildenhall2021nerf}) to synthesize a view of the surface being touched, which results in spatially aligned visual-tactile data (Fig.~\ref{fig:dataset_comp}).
We collect this data by mounting a touch sensor to a camera with commonly available materials (Fig.~\ref{fig:capturing_device}). To calibrate the pair of sensors, we take advantage of the fact that
popular vision-based touch sensors~\cite{johnson2009retrographic,johnson2011microgeometry,sferrazza2019design,lambeta2020digit} are built on ordinary cameras. The relative pose between the vision and tactile sensors can thus be estimated using traditional methods from multi-view geometry, such as camera resectioning~\cite{hartley2003multiple}.

We use this procedure to collect a large real-world dataset of aligned visual-tactile data. With this dataset, we train a diffusion model~\cite{sohl2015deep,rombach2021highresolution} to estimate touch at locations not directly probed by a sensor.
In contrast to the recent work of Zhong \etal~\cite{zhong2023touching}, which also estimates touch from 3D NeRF geometry, we create scene-scale reconstructions, we do not require robotic proprioception, and we use diffusion models~\cite{sohl2015deep}. This enables us to obtain tactile data at a much larger scale, and with considerably more diversity.  Unlike previous visual-tactile diffusion work~\cite{yang2023generating}, we condition the model on spatially aligned visual and depth information, enhancing the generated samples' quality and their usefulness in downstream applications.
After training, the diffusion model can be used to predict tactile information for novel positions in the scene. 
Analogous to quasi-dense stereo methods~\cite{lhuillier2005quasi,furukawa2009accurate}, the diffusion model effectively propagates sparse touch samples, obtained by probing, to other visually and structurally similar regions of the scene.

We evaluate our visual-tactile model's ability to accurately perform cross-modal translation using a variety of  quality metrics. We also apply it to several downstream tasks, including localizing a touch within a scene and understanding material properties of the touched area. Our experiments suggest:
\begin{itemize}[leftmargin=*,topsep=1pt, noitemsep]
    \item Touch signals can be localized in 3D space by exploiting multi-view geometry constraints between sight and touch.
    \item Estimated touch measurements from novel views are not only qualitatively accurate, but also beneficial on downstream tasks. %
    \item Cross-modal prediction models can accurately estimate touch from sight for natural scenes.
    \item Visually-acquired 3D scene geometry improves cross-modal prediction.
\end{itemize}

\mysection{Related Work}

\mypar{Visual-tactile datasets.}  Previous work has either used simulators~\cite{gao2021objectfolder,gao2022objectfolder} or robotic arms~\cite{calandra2017feeling,calandra2018moreta,li2019connecting,zhong2023touching,gao2023objectfolder} for data generation.
Our work is closely related to that of Zhong \etal~\cite{zhong2023touching}, which uses a NeRF and captured touch data to generate a tactile field for several small objects. They use the proprioception of an expensive robot to spatially align vision and touch. In contrast, we leverage the properties of the tactile sensor and novel view synthesis to use commonly available material (a smartphone and a selfie stick) to align vision and touch.
This enables the collection of a larger, scene-level, and more diverse dataset, on which we train a higher-capacity diffusion model (rather than a conditional GAN).
Like several previous works~\cite{burka2018instrumentation,yang2022touch}, we also collect scene-level data. In contrast to them, we spatially align the signals by registering them in a unified 3D representation, thereby increasing the prediction power of the visual-tactile generative model.

\mypar{Capturing multimodal 3D scenes.}
Our work is related to methods that capture 3D visual reconstructions of spaces using RGB-D data~\cite{silberman2012indoor,xiao2013sun3d,dai2017scannet,yeshwanth2023scannet++} and multimodal datasets of paired 3D vision and language~\cite{chen2020scanrefer,achlioptas2020referit3d,anderson2021sim}. Our work is also related to recent methods that localize objects in NeRFs using joint embeddings between images and language~\cite{kerr2023lerf} or by semantic segmentation~\cite{zhi2021place}. In contrast to language supervision, touch is tied to a precise position in a scene.

\mypar{3D touch sensing.} 
A variety of works have studied the close relationship between geometry and touch, motivating our use of geometry in imputing touch. Johnson \etal~\cite{johnson2009retrographic,johnson2011microgeometry} proposed vision-based touch sensing, and showed that highly accurate depth can be estimated from the touch sensor using photometric stereo. Other work has estimated object-scale 3D from touch~\cite{wang20183d}. By contrast, we combine sparse estimates of touch with quasi-dense tactile signals estimated using generative models.

\mypar{Cross-modal prediction of touch from sight.}
Recent work has trained generative models that predict touch from images. Li \etal~\cite{li2019connecting} used a GAN to predict touch for images of a robotic arm, while Gao \etal~\cite{gao2023objectfolder} applied them to objects collected on a turntable. Yang \etal~\cite{yang2023generating} used latent diffusion to predict touch from videos of humans touching objects. Our goal is different from these works: we want to predict touch signals that are spatially aligned with a visual signal, to exploit scene-specific information, and to use geometry. Thus, we use a different architecture and conditioning signal, and fit our model to examples from the same scenes at training and test time. Other work has learned joint embeddings between vision and touch~\cite{yuan2017connecting,lin2019learning,kerr2022ssvtp,yang2022touch, yang2024unitouch}.

\section{Method}

We collect visual and tactile examples from a scene and register them together with a 3D visual reconstruction to build a \vtfield. Specifically,  we capture a NeRF $\fieldv : (\pt, \view) \mapsto (\rgb, \sigma)$ that maps a 3D point $\pt = (x, y, z)$ and viewing direction $\view$ to its corresponding RGB color $\rgb$ and density $\sigma$~\cite{mildenhall2021nerf}.
We associate to the visual representation a {\em touch model} $\fieldt : \im_t \mapsto \touch$ that generates the tactile signal that one would obtain by touching at the center of the image $\im_t$.
In the following, we explain how to estimate $\fieldv$ and $\fieldt$ and put them into the same shared 3D space.

\subsection{Capturing vision and touch signals}

\mypar{Obtaining a visual 3D reconstruction.} We build the visual NeRF, $\fieldv$, closely following previous work~\cite{xiao2013sun3d,dai2017scannet}. A human data collector moves through a scene and records a video, covering as much of the space as possible. We then estimate camera pose using structure from motion~\cite{schoenberger2016sfm} and create a NeRF using off-the-shelf packages~\cite{tancik2023nerfstudio}. Additional details are provided in the supplement.

\mypar{Capturing and registering touch.} We simultaneously collect tactile and visual signals by mounting a touch sensor on a camera (Fig.~\ref{fig:capturing_device}), obtaining synchronized touch signals $\{\touch_i\}_{i=1}^N$ and video frames $\vidt$. We then estimate the pose of the video frames using off-the-shelf structure from motion methods~\cite{schoenberger2016sfm}, obtaining poses $\{p^v_i\}_{i=1}^N$. Finally, we use the calibration of the mount to obtain the poses $\{p^t_i\}_{i=1}^N$ of the tactile measurements with respect to the scene's global reference frame.
As a collection device, we mount an iPhone 14 Pro to one end of a camera rod, and a DIGIT~\cite{lambeta2020digit} touch sensor to the other end. Note that the devices can be replaced with any RGB-D camera and vision-based tactile sensor.

\mypar{Capturing setup calibration.} To find the relative pose between the camera and the touch sensor (Fig.~\ref{fig:capturing_device}),
we exploit the fact that arbitrary viewpoints can be synthesized from $\fieldv$, and that ubiquitous {\em vision-based} touch sensors are based on perspective cameras.
In these sensors, an elastomer gel is placed on the lens of a commodity camera, which is illuminated by colored lights. When the gel is pressed into an object, it deforms, and the camera records an image of the deformation; this image is used as the tactile signal. This design allows us to estimate the pose of the tactile sensor through multi-view constraints from {\em visual-tactile} correspondences: pixels in visual images and tactile images that are of the same physical point.

We start the calibration process by synthesizing novel views from $\fieldv$. The views are generated at the camera location $\{p^v_i\}_{i=1}^N$, but rotated $90^\circ$ on the $x$-axis. This is because the camera is approximately orthogonal to the touch sensor (see Fig.~\ref{fig:capturing_device}).
Then, we manually annotate corresponding pixels between the touch measurements and the generated frames (Fig.~\ref{fig:capturing_device}). To simplify and standardize this process, we place a braille board in each scene and probe it with the touch sensor. This will generate a distinctive touch signal that is easy to localize~\cite{higuera2023learning}.

We formulate the problem of estimating the six degrees of freedom relative pose $(\mathbf{R}, \mathbf{t})$ between the touch sensor and the generated frames as a resectioning problem~\cite{hartley2003multiple}. 
We use the estimated 3D structure from the NeRF $\fieldv$ to obtain 3D points $\{\pt_i\}_{i=1}^M$ for each of the annotated correspondences. 
Each point has a pixel position $\bu_i \in \mathcal{R}^2$ in the touch measurement. We find $(\mathbf{R}, \mathbf{t})$ by minimizing the reprojection error:
\vspace{-1pt}
\begin{equation}
\min_{{\mathbf R}, {\mathbf t}} \frac{1}{M}\sum_{i=1}^M 
\lVert \pi({\mathbf K}[\mathbf{R}\,\,|\,\,\mathbf{t}], \mathbf{X}_i) - \bu_i \rVert_1,
\end{equation}
where $\pi$ projects a 3D point using a given projection matrix, $\mathbf{K}$ are the known intrinsics of the tactile sensor's camera, and the point $\mathbf{X}_i$ is in the coordinate system of the generated vision frames. 
We perform the optimization on 6-15 annotated correspondences from the braille board. For robustness, we compute correspondences from multiple frames. 
We represent the rotation matrix using quaternions and optimize using nonlinear least-squares. Once we have $(\mathbf{R}, \mathbf{t})$ with respect to the generated frames, we can derive the relative pose between the camera and the touch sensor.

\subsection{Imputing the missing touch}
 \label{sec:estimating_dense_touch}

\begin{figure}[t]
    \centering
    \includegraphics[width=\linewidth]{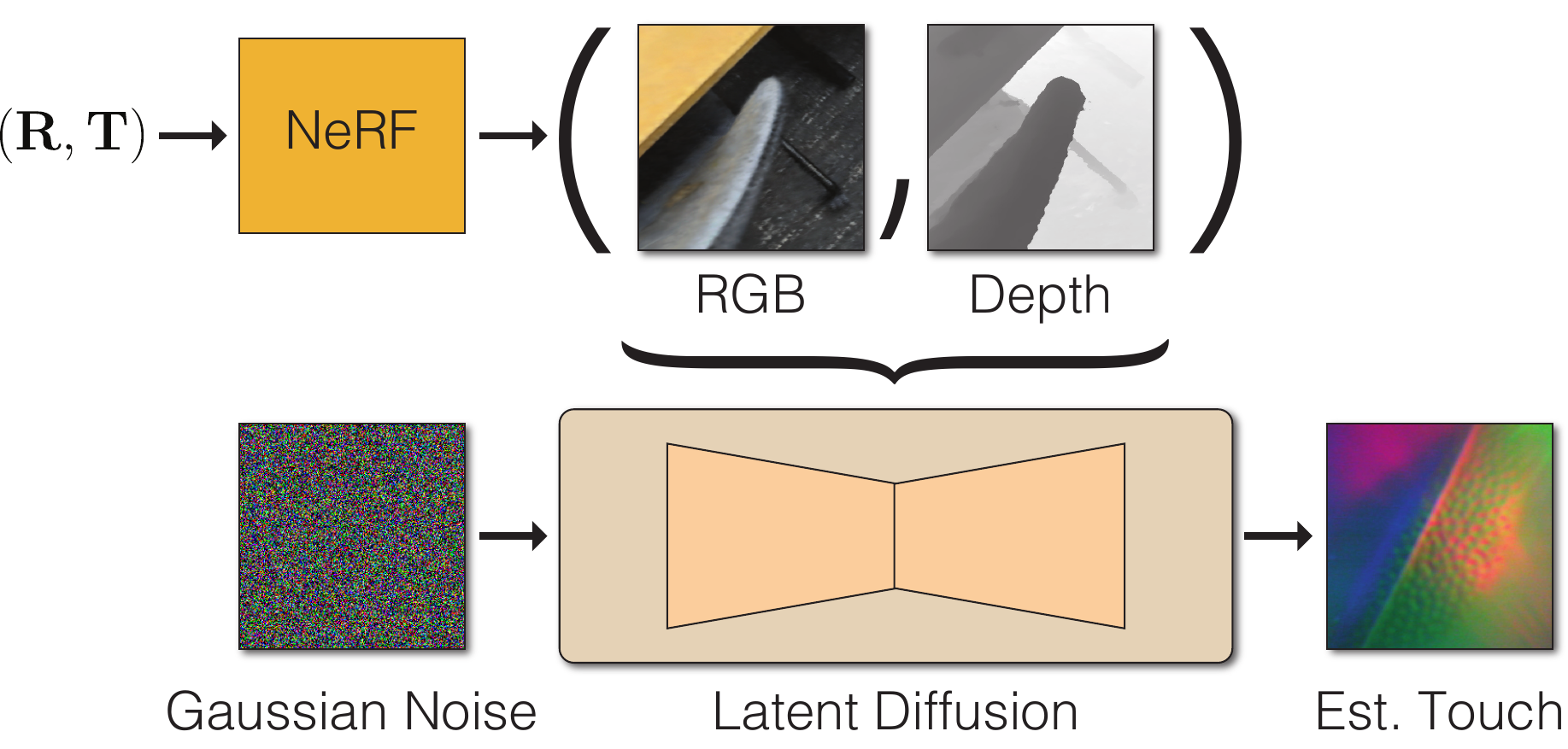}
    \caption{{\bf Touch estimation.} We estimate the tactile signal for a given touch sensor pose $(\mathbf{R}, \mathbf{t})$. To do this, we synthesize a viewpoint from the NeRF, along with a depth map. We use conditional latent diffusion to predict the tactile signal from these inputs.}\vspace{-3mm}
    \label{fig:method}
\end{figure}

We use a generative model to estimate the touch signal (represented as an image from a vision-based touch sensor) for other locations within the scene. %
Specifically, we train a diffusion model $p_{\touchparams}(\touch\mid\im, \depth,\background)$, where $\im$ and $\depth$ are images and depth maps extracted from $\fieldv$ (see Fig.~\ref{fig:method}).
We also pass as input to the diffusion model a {\em background} image captured by the touch sensor when it is not in contact with anything, denoted as $\background$. Although not  essential, we have observed that this additional input empirically improves the model's performance (\eg, Fig.~\ref{fig:teaser} the background provides the location of defects in the gel, which appear as black dots).
We train the model $p_{\touchparams}$ on our entire vision-touch dataset (Sec.~\ref{sec:dataset}).

The training of $p_{\touchparams}$ is divided into two stages.
In the first, we pre-train a cross-modal visual-tactile encoder with self-supervised contrastive learning on our dataset. This stage, initially proposed by~\cite{yang2023generating,higuera2023learning}, is equivalent to the self-supervised encoding pre-training that is common for image generation models~\cite{rombach2021highresolution}.
We use a ResNet-50~\cite{he2016deep} as the backbone for this contrastive model.

In the second stage, we use the contrastive model to generate the input for a conditional latent diffusion model, which is built upon Stable Diffusion~\cite{rombach2021highresolution}.
A frozen pretrained VQ-GAN~\cite{ding2021vq} is used to obtain the latent representation with a spatial dimension of $64\times 64$.
We start training the diffusion model from scratch and pre-train it on the task of unconditional tactile image generation on the YCB-Slide dataset~\cite{suresh2023midastouch}.
After this stage, we train the conditional generative model $p_{\touchparams}$ on our spatially aligned visual-tactile dataset, further fine-tuning the contrastive model end-to-end with the generation task.

At inference time, given a novel location in the 3D scene, we first render the visual signals $\hat{\im}$ and $\hat{\depth}$ from NeRF, and then estimate the touch signal $\hat{\touch}$ of the position using the diffusion model.

\section{A 3D Visual-Tactile Dataset}
\label{sec:dataset}
In the following, we show the details of the data collection process and statistics of our dataset.

\begin{figure*}[ht]
    \centering
    \includegraphics[width=\textwidth]{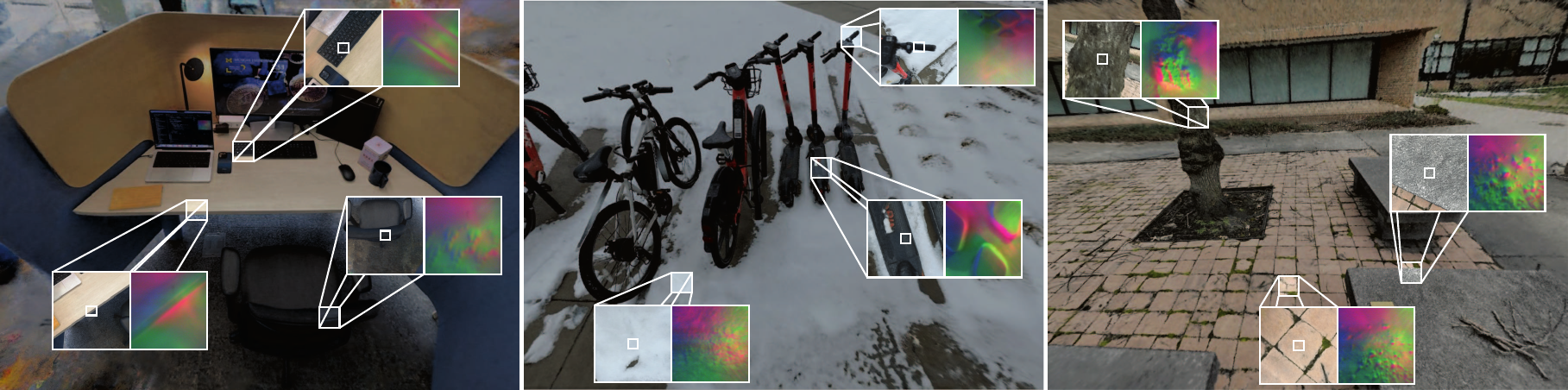}
    \caption{{\bf Representative examples from the captured dataset}. Our dataset is obtained from nine everyday scenes, such as offices, classrooms, and kitchens. We show three such scenes in the figure above, together with samples of spatially aligned visual and tactile data. In each scene, 1k to 2k tactile probes were collected, resulting in a total of 19.3k image pairs. The data encompasses diverse geometries (edges, surfaces, corners, etc.) and textures (plastic, clothes, snow, wood, etc.) of various materials. The collector systematically probed different objects, covering areas with distinct geometry and texture using different sensor poses.
    }
    \label{fig:dataset_example}
\end{figure*}

\subsection{Data Collection Procedure}

The data collection procedure is divided into two stages. 
First, we collect multiple views from the scene, capturing enough frames around the areas we plan to touch. During this stage, we collect approximately 500 frames.
Next, we collect synchronized visual and touch data, maximizing the geometry and texture being touched.
We then estimate the camera location of the vision frames collected in the previous two stages using off-the-shelf mapping tools~\cite{schoenberger2016sfm}.
After estimating the camera poses for the vision frames, the touch measurements' poses can be derived by using the mount calibration matrix.
More details about the pose estimation procedure can be found in the supplement.

Finally, we associate each touch sensor with a color image by translating the sensor poses upwards by $0.4$ meters and querying the NeRF with such poses. The field of view we use when querying the NeRF is $50^{\circ}$. This provides us with approximately 1,500 temporally aligned vision-touch image pairs per scene.
Note that this collection procedure is scalable since it does not require specific expertise or equipment and generates abundant scene-level samples.

\subsection{Dataset Statistics}
We collect our data in 13 ordinary scenes including two offices, a workroom, a conference room, a corridor, a tabletop, a corridor, a lounge, a room with various clothes and four outdoor scenes with interesting materials. Typically, we collect 1k to 2k tactile probes in each scene, resulting in a total of 19.3k image pairs in the dataset.

Some representative samples from the collected dataset are shown in Fig.~\ref{fig:dataset_example}. Our data includes a large variety of geometry (edges, surfaces, corners, \etc) and texture (plastic, clothes, snow, wood, \etc) of different materials in the scene.
During capturing process, the collector will try to thoroughly probe various objects and cover the interesting areas with more distinguishable geometry and texture with different sensor poses.
To the best of our knowledge, our dataset is the first dataset that captures full, scene-scale spatially aligned vision-touch image pairs. 
We provide more details about the dataset in the supplement. %

\section{Experiments}
Leveraging the spatially aligned image and touch pairs from our dataset, we first conduct experiments on dense touch estimation. We then show the effectiveness of both the aligned data pairs and the synthesized touch signals by conducting tactile localization and material classification as two downstream tasks.

\subsection{Implementation Details}
\label{sec:implementation}

\mypar{NeRF.} We use the Nerfacto method from Nerfstudio~\cite{tancik2023nerfstudio}. 
For each scene, we utilize approximately 2,000 images as training set, which thoroughly cover the scene from various view points.
We train the network with a base learning rate of $1 \times 10^{-2}$ using Adam~\cite{kingma2014adam} optimizer for 200,000 steps on a single NVIDIA RTX 2080 Ti GPU to achieve optimal performance.

\mypar{Visual-tactile contrastive model.} 
Following prior works~\cite{kerr2022learning,yang2023generating}, we leverage contrastive learning methods to train a ResNet-50~\cite{he2016deep} as visual encoder. The visual and tactile encoders share the same architecture but have different weights. We encode visual and tactile data into latent vectors in the resulting shared representation space. We set the dimension of the latent vectors to 32. 
Similar to CLIP~\cite{radford2021learning}, the model is trained on InfoNCE loss obtained from the pairwise dot products of the latent vectors.
We train the model for 20 epochs by Adam~\cite{kingma2014adam} optimizer with a learning rate of $10^{-4}$ and batch size of 256 on 4 NVIDIA RTX 2080 Ti GPUs.

\mypar{Visual-tactile generative model.}
Our implementation of the diffusion model closely follows Stable Diffusion~\cite{rombach2022high}, with the difference that we use a ResNet-50 to generate the visual encoding from RGB-D images for conditioning. 
Specifically, we also add the RGB-D images rendered from the tactile sensors' poses into the conditioning, which we refer to in Sec.~\ref{sec:dense_touch} as multiscale conditioning. 
The model is optimized for 30 epochs by Adam~\cite{kingma2014adam} optimizer with a base learning rate of $10^{-5}$. The learning rate is scaled by $\text{gpu number} \times \text{batch size}$. We train the model with batch size of 48 on 4 NVIDIA A40 GPUs. 
At inference time, the model conducts 200 steps of denoising process with a $7.5$ guidance scale. 
Following prior cross-modal synthesis work~\cite{ramesh2021zero}, we use reranking to improve the prediction quality. We obtain 16 samples from the diffusion model for every instance and re-rank the samples with our pretrained contrastive model. The sample with highest similarity is the final prediction.

\subsection{Dense Touch Estimation}
\label{sec:dense_touch}

\begin{figure*}[t]
    \centering
    \includegraphics[width=\textwidth]{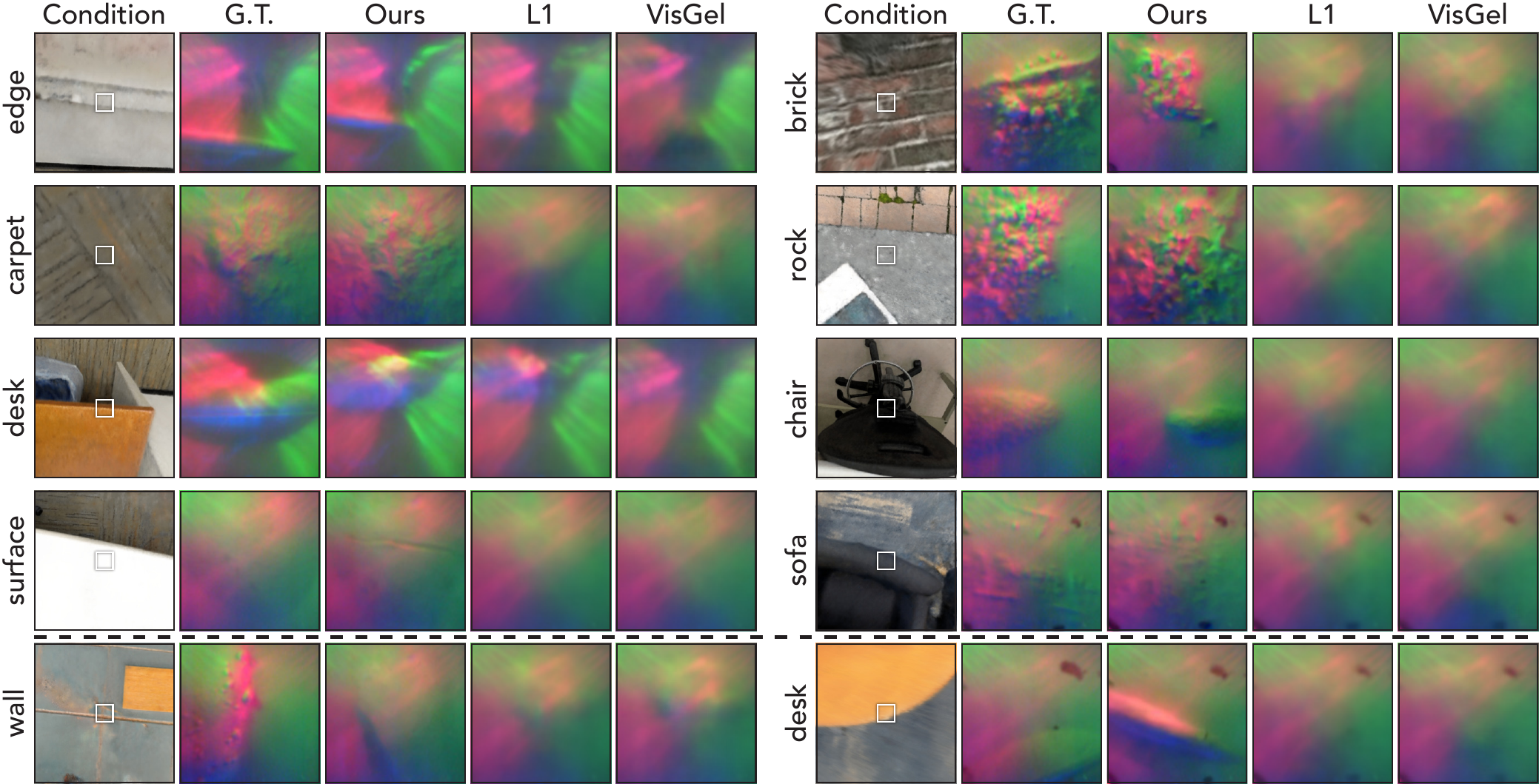}\vspace{-1mm}
    \caption{{\bf Qualitative touch estimation results}. Each model is conditioned on the RGB image and depth map rendered from the NeRF (left). The white box  indicates the tactile sensor's approximate field of view (which is much smaller than the full conditional image). The G.T. column shows the ground truth touch images measured from a DIGIT sensor. L1 and VisGel often generate blurry textures and inaccurate geometry. By contrast, our model better captures the features of the tactile image, \eg, the rock's  microgeometry and complex textures and shapes of furniture. The last row shows two failure cases of our model. In both examples, our model generates a touch image that is geometrically misaligned with the ground truth. All of the examples shown here are at least $10$cm away from any training sample.}
    \label{fig:estimation_example}
\end{figure*}

\paragraph{Experimental setup.} We now evaluate the diffusion model's ability to generate touch images. To reduce overlap between the training and test set, we first split the frames into sequences temporally (following previous work~\cite{yang2022touch}). We split them into sequences of 50 touch samples, then divide these sequences into train/validation/test with a ratio of 8/1/1.
We evaluate the generated samples on Frechet Inception Distance (FID), a standard evaluation metric for cross-modal generation~\cite{yang2022touch}. We also include Peak Signal to Noise Ratio (PSNR) and Structural Similarity (SSIM), though we note that these metrics are highly sensitive to spatial position of the generated content, and can be optimized by models that minimize simple pixelwise losses~\cite{heusel2017gans}. We also include CVTP metric proposed by prior work~\cite{yang2023generating}, which measures the similarity between visual and tactile embeddings of a contrastive model, analogous to CLIP~\cite{radford2021learning} score.
We compare against two baselines: \emph{VisGel}, the approach from Li et.~\cite{li2019connecting}, which trains a GAN for touch generation, and \emph{L1}, a model with the same architecture of VisGel but trained to minimize an L1 loss in pixel space. %

\mypar{Results.}
As is shown in Table~\ref{tab:results}, our approach performs much better on the high-level metrics, with up to 4x lower FID and 80x higher CVTP.
This indicates that our proposed diffusion model captures the distribution and characteristics of the real tactile data more effectively.
On the low-level metrics (PSNR and SSIM), all methods are comparable. In particular, the L1 model slightly outperforms the other methods since the loss it is trained on is highly correlated with low-level, pixel-wise metrics.
Fig.~\ref{fig:estimation_example} qualitatively compares samples from the different models.
Indeed, our generated samples exhibit enhanced details in micro-geometry of fabrics and richer textures, including snow, wood and carpeting.
However, all methods fail on fine details that are barely visible in the image, such as the tree bark.

\begin{table}[t]
\small
    \centering
    \begin{tabularx}{\linewidth}{@{}Xcccc@{}}
    \toprule
        Model & PSNR $\uparrow$ & SSIM $\uparrow$& FID $\downarrow$& CVTP $\uparrow$\\
    \midrule
        L1 & $\textbf{24.34}$ & $\textbf{0.82}$ & $97.05$ &$0.01$\\
        VisGel~\cite{li2019connecting} & $23.66$ & $0.81$ & $130.22$ &$0.03$\\
        Ours & $22.84$ & $0.72$ & $\textbf{28.97}$ & $\textbf{0.80}$\\
    \bottomrule
    \end{tabularx}
    \vspace{-2mm}
    \caption{{\bf Quantitative results on touch estimation for novel views.} While comparable on low-level metrics with the baselines, our approach captures the characteristics of the real tactile data more effectively, resulting in a lower FID score.}
    \label{tab:results} \vspace{-5mm}
\end{table}

\mypar{Ablation study.} We evaluate the importance of the main components of our proposed touch generation approach (Table~\ref{tab:ablation}). %
Removing the conditioning on the RGB image results in the most prominent performance drop. This is expected since RGB image uniquely determines the fine-grained details of a tactile image.
Removing depth image or contrastive pretraining has small effect on CVTP but results in a drop on FID.
Contrastive re-ranking largely improves CVTP, indicating the necessity of obtaining multiple samples from the diffusion model.
We also find that multiscale conditioning provide a small benefit on FID and CVTP.

\begin{table}[t]
    \centering
    \small
    \setlength\tabcolsep{2pt}
    \begin{tabularx}{\linewidth}{Xlcccc@{}}
    \toprule
      Model variation   & PSNR $\uparrow$ & SSIM $\uparrow$& FID $\downarrow$& CVTP $\uparrow$\\
    \midrule
        Full & $22.84$ & $\textbf{0.72}$ & $\textbf{28.97}$ & $\textbf{0.80}$\\
        No RGB conditioning & $22.13$ & $0.70$ & $34.31$ & $0.76$\\
        No depth conditioning & $22.57$ &  $0.71$ & $33.16$ &$\textbf{0.80}$\\
        No contrastive pretraining & $22.82$ & $0.71$ & $32.98$ & $0.79$\\
        No re-ranking & $22.92$ & $\textbf{0.72}$ & $29.46$ &$0.61$\\
        No multiscale & $\textbf{23.19}$ & $\textbf{0.72}$ & $30.89$ & $0.77$\\
    \bottomrule
    \end{tabularx}
    \caption{{\bf Ablation study.} Since the fine-grained details of touch images can be determined from a RGB image, removing conditioning on the latter results in the largest performance drops. Re-ranking has notable impact on CVTP, indicating the necessity of obtaining multiple samples from the diffusion model.}
    \vspace{-8mm}
    \label{tab:ablation}
\end{table}

\subsection{Downstream Task I: Tactile Localization}
\begin{figure*}[t]
    \centering
    \includegraphics[width=\textwidth]{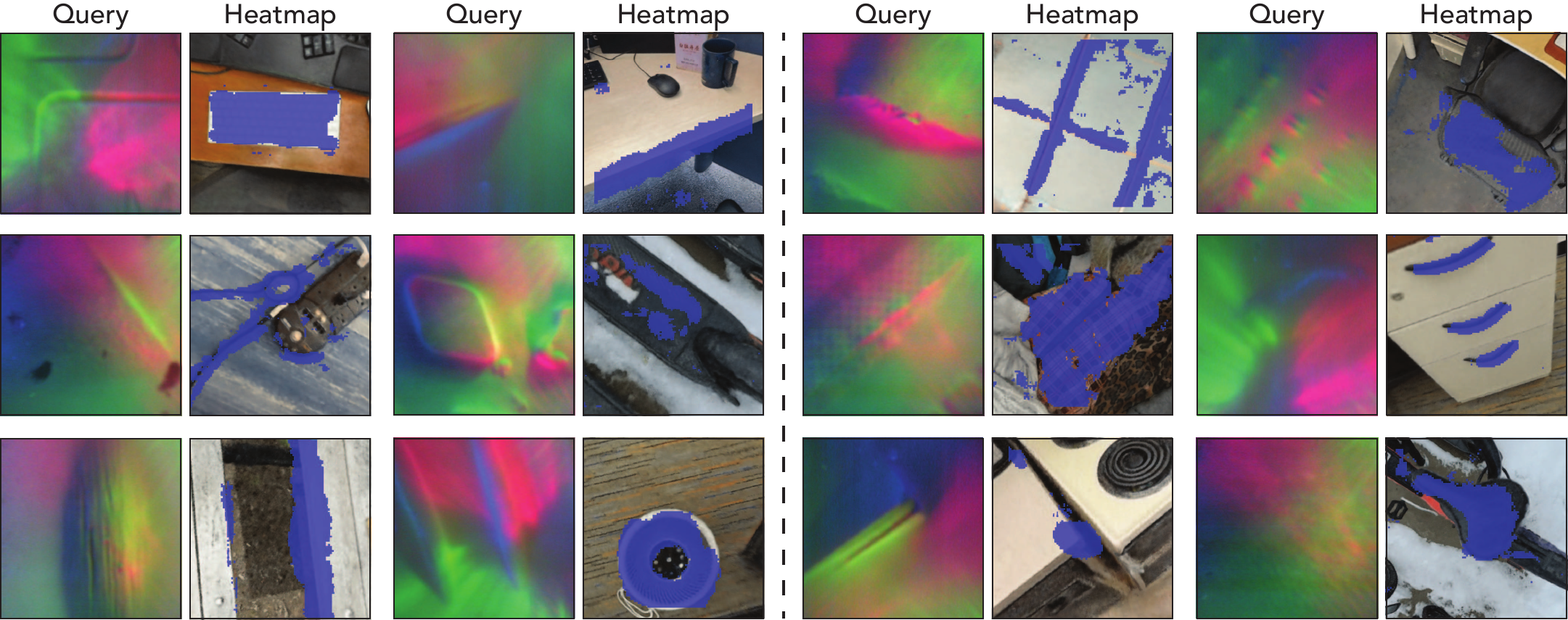}\vspace{-1mm}
    \caption{\textbf{Tactile localization heatmaps}. Given a tactile query image, the heatmap shows the image patches with a higher affinity to this tactile signal, as measured by a contrastive model trained on our dataset. We use a sliding window and compare each extracted patch with the touch signal. In each case, the center patch is the true position. Our model successfully captures the correlation between the two signals. This enables it to localize a variety of touch signals,  including fine-grained geometry, \eg, a cable or a keyboard, various types of corners and edges, and large uniform regions, such as a clothing. This ability enables our diffusion model to effectively propagate sparse touch samples to other visually and structurally similar regions of the scene.} %
    \label{fig:heatmap} \vspace{-3mm}
\end{figure*}

To help understand the quality of the captured TaRFs, we evaluate the performance of the contrastive model (used for conditioning our diffusion model) on the task of tactile localization. 
Given a tactile signal, our goal is to find the corresponding regions in a 2D image or in a 3D scene that are associated with it, \ie, we ask the question: \emph{what part of this image/scene feel like this?} We perform the following evaluations on the test set of our dataset. Note that we run no task-specific training.
\vspace{-5mm}

\paragraph{2D Localization.} To determine which part of an image are associated with a given tactile measurement, we follow the same setup of SSVTP~\cite{kerr2022ssvtp}. We first split the image into patches and compute their embedding. Then, we generate the tactile embedding of the input touch image. Finally, we compute the pairwise similarities between the tactile and visual embeddings, which we plot as a heatmap.
 As we can see in Fig.~\ref{fig:heatmap}, our constrastive encoder can successfully capture the correlations between the visual and tactile data. For instance, the tactile embeddings of edges are associated to edges of similar shape in the visual image. Note that the majority of tactile embeddings are highly ambiguous: all edges with a similar geometry \emph{feel} the same.

\paragraph{3D Localization.} In 3D, the association of an image to tactile measurements becomes less ambiguous. Indeed, since tactile-visual samples are rotation-dependent, objects with similar shapes but different orientations will generate different tactile measurements. Lifting the task to 3D still does not remove all ambiguities (for example, each side of a rectangular table cannot be precisely localized). Nonetheless, we believe it to be a good fit for a quantitative evaluation since it's rare for two ambiguous parts of the scene to be touched with \emph{exactly} the same orientation.

We use the following experimental setup for 3D localization. Given a tactile image as a query, we compute its distance in embedding space to all visual test images from the same scene. Note that all test images are associated with a 3D location. We define as ground-truth correspondences all test images at a distance of at most $r$ from the 3D location of the test sample. We vary $r$ to account for local ambiguities. As typical in the retrieval literature, we benchmark the performance with metric mean Average Precision (mAP).

We consider three baselines: (1) \emph{chance}, which randomly selects corresponding samples; (2) \emph{real}, which uses the contrastive model trained on our dataset; and (3) \emph{real + estimated}, which trains the contrastive model on both dataset samples and a set of synthetic samples generated via the scenes' NeRF and our touch generation model. Specifically, we render a new image and corresponding touch by interpolating the position of two consecutive frames in the training dataset. This results in a training dataset for the contrastive model that is twice as large.

The results, presented in Table~\ref{tab:tactile_localization}, demonstrate the performance benefit of employing both real and synthetic tactile pairs. Combining synthetic tactile images with the original pairs achieves highest performance on all distance thresholds.
Overall, this indicates that touch measurements from novel views are not only qualitatively accurate, but also beneficial for this downstream task.

\begin{table}[t]
{
\small
    \centering
    \label{my-label}
    \begin{tabularx}{\linewidth}{@{}Xcccccc@{}}
    \toprule
     & \multicolumn{5}{c}{$r (m)$} \\ \cmidrule(l){2-6} 
    Dataset & $0.001$ & $0.005$ & $0.01$ & $0.05$ & $0.1$ \\ \midrule
    Chance  & $3.55$ & $6.82$ & $10.25$ & $18.26$ & $21.33$\\
    Real  & $12.10$ & $22.93$ & $32.10$ & $50.30$ & $57.15$\\
    Real + Est. & $\textbf{14.92}$ & $\textbf{26.69}$ & $\textbf{36.17}$ & $\textbf{53.62}$ & $\textbf{60.61}$\\
    \bottomrule
    \end{tabularx}
    \vspace{-1mm}
    \caption{\textbf{Quantitative results on 3D tactile localization}. 
    We evaluate using mean Average Precision (mAP) as a metric. Training the contrastive model on our dataset of visually aligned real samples together with estimated samples from new locations in the scene results in the highest performance.}
    \label{tab:tactile_localization} \vspace{-3mm}
}
    
\end{table}

\subsection{Downstream Task II: Material Classification}

We investigate the efficacy of our visual-tactile dataset for understanding material properties, focusing on the task of material classification.  We follow the formulation by Yang \etal~\cite{yang2022touch}, which consists of three subtasks: (i) material classification, requiring the distinction of materials among 20 possible classes; (ii) softness classification, a binary problem dividing materials as either hard or soft; and (iii) hardness classification, which requires the classification of materials as either rough or smooth.

We follow the same experimental procedure of~\cite{yang2022touch}: we pretrain a contrastive model on a dataset and perform linear probing on the sub-tasks' training set.
Our experiments only vary the pretraining dataset, leaving all architectural choices and hyperparameters the same. We compare against four baselines. A random classifier (\emph{chance}); the ObjectFolder 2.0 dataset~\cite{gao2022objectfolder}; the VisGel dataset~\cite{li2019connecting}; and the Touch and Go dataset~\cite{yang2022touch}. Note that the touch sensor used in the test data (GelSight) differs from the one used in our dataset (DIGIT). Therefore, we use for pretraining a combination of our dataset and Touch and Go. To ensure a fair comparison, we also compare to the combination of each dataset and Touch and Go.

The findings from this evaluation, as shown in Table~\ref{tab:material}, suggest that our data improves the effectiveness of the contrastive pretraining objective, even though our data is from a {different} distribution. %
Moreover, we find that adding estimated touch probes for pretraining results in a higher performance on all the three tasks, especially the smoothness classification. This indicates that not only does our dataset covers a wide range of materials but also our diffusion model captures the distinguishable and useful patterns of different materials.

\begin{table}[t]
{\small
    \centering
    \begin{tabularx}{\linewidth}{X c c c}
    \toprule
        Dataset & Material & \begin{tabular}[c]{@{}c@{}}Hard/\\Soft\end{tabular}& \begin{tabular}[c]{@{}c@{}}Rough/\\Smooth\end{tabular}\\
    \midrule
        Chance &$18.6$&$66.1$&$56.3$\\
        ObjectFolder 2.0~\cite{gao2022objectfolder} &$36.2$&$72.0$&$69.0$\\
        VisGel~\cite{li2019connecting} &$39.1$&$69.4$&$70.4$\\
        \cmidrule{1-4}
        Touch and Go~\cite{yang2022touch} &$54.7$&$77.3$&$79.4$\\
    ~~~ + ObjectFolder 2.0~\cite{gao2022objectfolder} &$54.6$&$87.3$&$84.8$\\
    ~~~ + VisGel~\cite{li2019connecting} &$53.1$&$86.7$&$83.6$\\
    ~~~ + Ours$^*$ (Real) & $57.6$&$88.4$&$81.7$ \\ 
    ~~~ + Ours$^*$ (Real + Estimated) & $\textbf{59.0}$ & $\textbf{88.7}$ & $\textbf{86.1}$\\
    \bottomrule %
    \end{tabularx}
    \vspace{-1.25mm}
    \caption{{\bf Material classification}. We show the downstream material recognition accuracy of models pre-trained on different datasets. The final rows show the performance when combining different datasets with {\em Touch and Go}~\cite{yang2022touch}. $^*$ The task-specific training and testing datasets for this task are collected with a GelSight sensor. {We note that our data comes from a different distribution, since it is collected with a DIGIT sensor~\cite{lambeta2020digit}.}\vspace{-4mm}}
    \label{tab:material}}
\end{table}

\vspace{-2mm}
\section{Conclusion}

In this work, we present the \vtfield, a scene representation that brings vision and touch into a shared 3D space.
This representation enables the generation of touch probes for novel scene locations.
To build this representation, we collect the largest dataset of spatially aligned vision and touch probes.%
We study the utility of both the representation and the dataset in a series of qualitative and quantitative experiments and on two downstream tasks: 3D touch localization and material recognition.
Overall, our work makes the first step towards giving current scene representation techniques an understanding of not only how things look, but also how they \emph{feel}.
This capability could be critical in several applications ranging from robotics to the creation of virtual worlds that look and feel like the real world.

\mypar{Limitations.}
Since the touch sensor is based on a highly zoomed-in camera, small (centimeter-scale) errors in SfM or visual-tactile registration can lead to misalignments of several pixels between the views of the NeRF and the touch samples, which can be seen in our TaRFs. Another limitation of the proposed representation is the assumption that the scene's coarse-scale structure does not change when it is touched, an assumption that may be violated for some inelastic surfaces. 

\mypar{Acknowledgements.} We thank Jeongsoo Park, Ayush Shrivastava, Daniel Geng, Ziyang Chen, Zihao Wei, Zixuan Pan, Chao Feng, Chris Rockwell, Gaurav Kaul and the reviewers for the valuable discussion and feedback. This work was supported by  an NSF CAREER Award \#2339071, a Sony Research Award, the DARPA Machine Common Sense program, and ONR MURI award N00014-21-1-2801.

{
    \small
    \bibliographystyle{ieeenat_fullname}
    \bibliography{main}
}

\end{document}